\documentclass[10pt,letterpaper,twocolumn]{article}
\usepackage[latin9]{inputenc}
\usepackage{amsmath}
\usepackage{amssymb}
\usepackage{graphicx}
\usepackage[unicode=true,
 bookmarks=false,
 breaklinks=false,pdfborder={0 0 1},backref=section,colorlinks=false]
 {hyperref}
\hypersetup{
 pagebackref,breaklinks,colorlinks}

\makeatletter

\pdfpageheight\paperheight
\pdfpagewidth\paperwidth

\usepackage{cvpr}              

\usepackage{graphicx}
\usepackage[accsupp]{axessibility}

\usepackage[capitalize]{cleveref}
\crefname{section}{Sec.}{Secs.}
\Crefname{section}{Section}{Sections}
\Crefname{table}{Table}{Tables}
\crefname{table}{Tab.}{Tabs.}

\AtBeginDocument{%

\let\ref\cref
}

\usepackage{enumitem}
\setlist[itemize]{leftmargin=*}
\usepackage{balance}
\usepackage{microtype}
\newcommand\scalemath[2]{\scalebox{#1}{\mbox{\ensuremath{\displaystyle #2}}}}

\def\cvprPaperID{3} 
\def\confName{CVPR}
\def\confYear{2022}

\@ifundefined{showcaptionsetup}{}{%
 \PassOptionsToPackage{caption=false}{subfig}}
\usepackage{subfig}
\makeatother

\begin{document}
\title{Persistent-Transient Duality in Human Behavior Modeling}
\author{Hung Tran, Vuong Le, Svetha Venkatesh, Truyen Tran\\
Applied AI Institute, Deakin University, Geelong, Australia \\
 \texttt{\small{}\{tduy,vuong.le,svetha.venkatesh,truyen.tran\}@deakin.edu.au}}
\maketitle
\begin{abstract}
We propose to model the persistent-transient duality in human behavior
using a parent-child multi-channel neural network, which features
a parent persistent channel that manages the global dynamics and children
transient channels that are initiated and terminated on-demand to
handle detailed interactive actions. The short-lived transient sessions
are managed by a proposed Transient Switch. The neural framework is
trained to discover the structure of the duality automatically. Our
model shows superior performances in human-object interaction motion
prediction.

\end{abstract}
\global\long\def\ModelName{\text{Egocentric Graph Interaction Network}}%
\global\long\def\Model{\text{EGIN}}%
\global\long\def\OnePart{\text{mechanism}}%
\global\long\def\TheoryMain{\text{Persistent process}}%
\global\long\def\TheoryLoc{\text{Transient process}}%

\global\long\def\ImpMain{\text{Persistent channel}}%
\global\long\def\ImpLoc{\text{\text{Transient channel}}}%

\global\long\def\SwitchModule{\text{\text{Switch module}}}%

\section{Introduction}

\begin{figure}
\begin{centering}
\includegraphics[width=0.95\columnwidth]{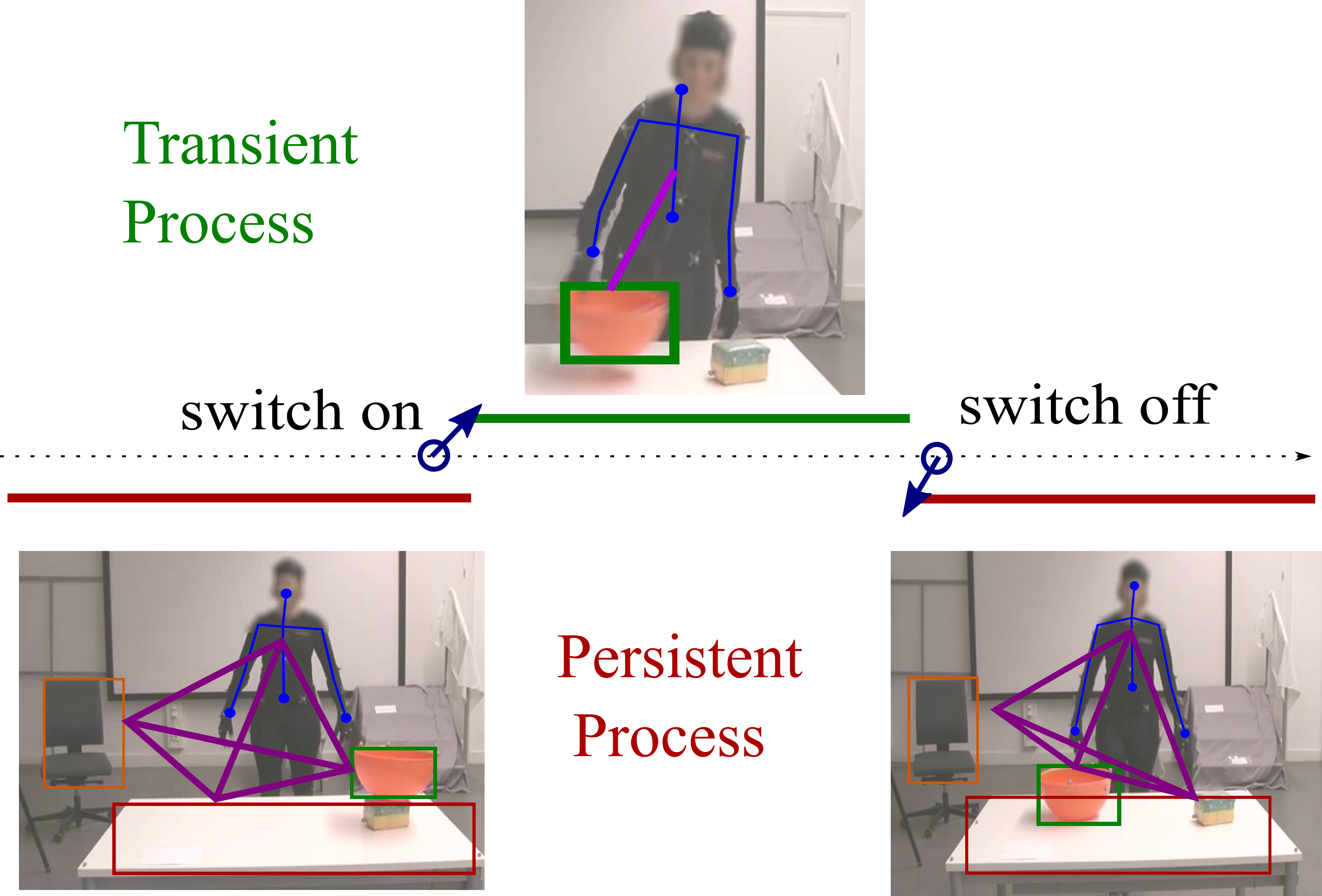}
\par\end{centering}
\caption{We model the \emph{Persistent-transient duality} of human behaviors.
In the default \emph{Persistent process} (lower row), the human subjects
consider the whole scene from the global view with dense relational
connections (purple links). For an interactive action, they switch
on a \emph{Transient process} (upper row) that zooms into a small
set of local objects that are directly interacted. When done with
such task, they switch back to the default Persistent mode. \label{fig:intro_demonstration}}
\end{figure}

Human behavior is highly contextualized, reacting to the rapid changes
in the situation. This fast-changing nature requires a model to adapt
quickly in structure, representation, and inference mechanism to follow
the true patterns of the behavior. Such requirement is critical in
the problem of human-object interaction (HOI) motion prediction, where
the subjects follow an overall plan but occasionally deviate from
it to solve emerging tasks\cite{baker2014fast}. 

Motion models on this task reflect the changing situations by gradually
adapting their relational structure. However, with a fixed inference
mechanism, they cannot account for the discrete switching between
distinctive mechanisms and fail to keep up with the movement patterns.
For example, when the human subject deviates from the overall path
to interact with an object, these models will continue to consider
the interacted object as an equal member of the scene and miss its
importance as the action's direct recipient.

We address this limitation by factorizing the human behaviors into
two processes: a \emph{slow-changing persistent process}, which maintains
a continuous default dynamic, and a \emph{fast-changing transient
process}, which has an adaptive life-cycle and a personalized structure
that reflects the human's perspective in emerging events (See \ref{fig:intro_demonstration}).
The two processes are modeled into a parent-child multi-channel neural
network called \emph{Persistent-Transient Duality}. The \emph{Persistent
channel} is a recurrent relational network operating on the global
scene spatially and throughout the session temporally. The \emph{Transient
channels} instead have a contextualized graphical structure constructed
on the spot whenever the human subjects shift the priority toward
interacting with other entities. The life cycles of these channels
are managed by a neural \emph{Transient Switch}, which can learn to
anticipate when a Transient channel will be needed and trigger it
in time. 

Our model establishes the SOTA performance on the HOI subset of the
KIT Whole-Body Human Motion Database.

\section{Method}

\begin{figure*}
\begin{centering}
\includegraphics[width=0.9\textwidth]{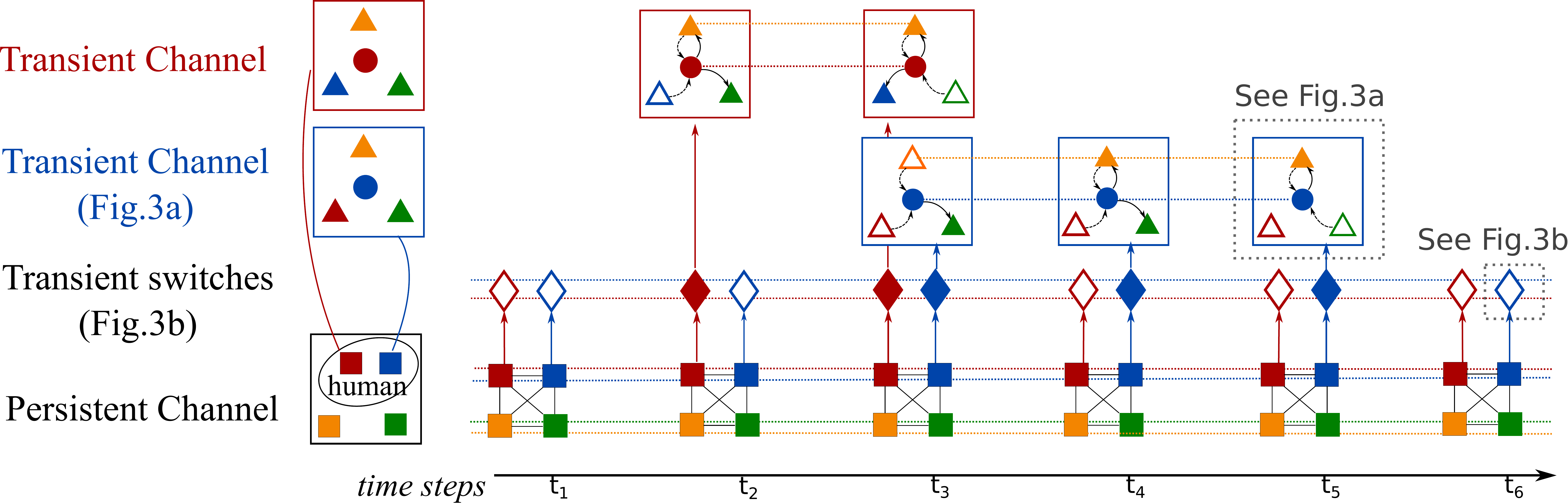}
\par\end{centering}
\caption{The architecture of \emph{Persistent-transient Duality }Networks (PTD).
The \emph{persistent channel }contains fully-connected recurrent graph
networks of all entities (squares) similar to current SoTA works.
We introduce the new \emph{Transient channel} to zoom into the local
context of each human (circles) when they interact with surrounding
entities (triangles). The transient channels are initialized and terminated
on-demand, controlled by the neural \emph{Transient Switches} (diamonds).
\label{fig:overal_architecture}}
\end{figure*}

\subsection{Preliminaries}

We consider the problem of modeling the sequential behaviors of $N$
entities in a dynamic system, where each entity $i$ is represented
by a class label $c_{i}$ and sequential features $X_{i}=\left\{ x_{i}^{t}\right\} _{t=1}^{T}$.
After observing $T$ steps, we predict the features in the next $L$
steps, $Y_{i}=\left\{ y_{i}^{t}\right\} _{t=T+1}^{T+L+1}$. The entity
classes include the human and object ($c_{i}$=''human''/''object''),
which decide the entity's feature spaces and behaviors.

In this paper, we use a customized attention function defined over
the query $q\in R^{d}$ and the identical key/value pairs $V=\{v_{j}\}_{j=1}^{N}\in R^{d\times N}$:

$$\scalemath{0.95}{\text{Attn}\left(q,V\right)=\sigma\left(\sum_{j=1}^{N}\text{softmax}_{j}\left(\boldsymbol{W}_{\alpha}^{T}\left[\boldsymbol{W}_{q}q;\boldsymbol{W}_{v}v_{j}\right]\right)\boldsymbol{W}_{v}v_{j}\right)} $$

where $\left[\cdot;\cdot\right]$ is the concatenation, $\sigma$
is a non-linear activation function, $\boldsymbol{W}_{v}$, $\boldsymbol{W}_{q}$,
and $\boldsymbol{W}_{\alpha}$ are learnable weights.

\subsection{The Persistent-Transient Duality }

We model the persistent-transient duality of human behavior by a hierarchical
neural network called \emph{Persistent-Transient Duality Networks
(PTD)} (See \ref{fig:overal_architecture}). The network has three
main components: The Persistent Channel, the Transient Channels, and
the Transient Switch.

\subsection{Persistent Channel\label{subsec:Persistent-Channel}}

The Persistent channel oversees the global view of the scene, including
the humans and other entities. It has the form of a recurrent relational
network where entities interact in the spatio-temporal space spanned
by the video. The temporal evolution of each entity is modeled as:

\vspace{-1em}

\begin{equation}
h_{i}^{\mathcal{P},t}=\text{RNN}_{c_{i}}\left(\left[z_{i}^{\mathcal{P},t},m_{i,\mathcal{T}\rightarrow\mathcal{P}}^{t}\right],h_{i}^{\mathcal{P},t-1}\right),\label{eq:persistent-rnn}
\end{equation}
where $\text{RNN}_{c_{i}}$ is a recurrent unit that corresponds to
the class of the $i^{\text{th}}$ entity, maintaining hidden states
$h_{i}^{\mathcal{P},t-1}$ and consuming input vector $z_{i}^{\mathcal{P},t}$.
This input is formed as $z_{i}^{\mathcal{P},t}=\left[x_{i}^{t};m_{i}^{t}\right]$,
where $x_{i}^{t}$ is the entity's feature and $m_{i}^{t}$ is the
message aggregated from the entity's neighbor through attention, $m_{i}^{t}=\text{Attn}\left(u_{i}^{t},\left\{ u_{j}^{t}\right\} _{j\neq i}\right)$
with $u\_^{t}=\left[x\_^{t};h\_^{\mathcal{P},t-1}\right]$.

The unit also uses an optional transient-persistent message $m_{i,\mathcal{T}\rightarrow\mathcal{P}}^{t}$
which is non-zero if a corresponding \emph{transient process} (\ref{subsec:transient_process})
is currently active. 

The Persistent channel generates two outputs from its hidden state:
the future prediction $\hat{y}_{i}^{\mathcal{P},t}$ and the message
$m_{i,\mathcal{P}\rightarrow\mathcal{T}}^{t}$ to the Transient channel:
\begin{equation}
\hat{y}_{i}^{\mathcal{P},t}=\text{MLP}\left(h_{i}^{\mathcal{P},t}\right),\hspace{0.3cm}m_{i,\mathcal{P}\rightarrow\mathcal{T}}^{t}=\text{MLP}\left(h_{i}^{\mathcal{P},t}\right).\label{eq:persistent_output}
\end{equation}
The channel's prediction output $\hat{y}_{i}^{\mathcal{P},t}$ are
combined with those from the Transient channel as detailed in \ref{subsec:Future-prediction}.

This persistent channel is equivalent in modeling with the major state-of-the-art
HOI-M by using relational recurrent models \cite{corona2020context}.
Our key novelty is the consideration of the second side of the duality
- the Transient process, presented in the next section.

\subsection{Transient Channel \label{subsec:transient_process}}

Within the persistent-transient duality, the Transient process allows
the model to zoom in at relevant context and take the local viewpoint
of the human entity when it starts to interact with objects. This
human-specific process is implemented by a Transient channel. The
egocentric property of this channel separates it from the global view
of its parent persistent channel and reflects in three aspects of
\emph{feature representation}, \emph{computational structure}, and
\emph{inference logic}.

\paragraph{The egocentric graph structure}

reflects the relations between the active subject and the surrounding
passive entities. We define the \emph{Transient graph} for a human
entity of index $i$ at time $t$ to be $\mathcal{G}_{i}^{t}=\left(\mathcal{V}_{i}^{t},\mathcal{E}_{i}^{t}\right)$.
For a particular human, subscript $i$ will be omitted for conciseness. 

The egocentric characteristic of $\mathcal{G}^{t}$ reflects in its
star-like structure: the nodes $\mathcal{V}^{t}$ includes a single
\emph{center node $r$} for the considering human, and \emph{leaf
nodes} of indices $\left\{ l\right\} _{l\neq r}$ for other entities.
The dynamic edges $\mathcal{E}^{t}$ connect the center with the leaves
in two directions: \emph{inward edges }$e_{l\rightarrow r}^{t}$ reflect
which objects the human pay attention to, and the \emph{outward edges}
$e_{r\rightarrow l}^{t}$ represents the objects are being manipulated
by the human. These edges are determined at each time step by thresholding
the center-leaf distances $d_{lr}^{t}$, making the graph's topology
evolve within one single Transient session.

\begin{figure}
\subfloat[Transient Channel\label{fig:transient-channel}]{\begin{centering}
\includegraphics[width=0.47\columnwidth]{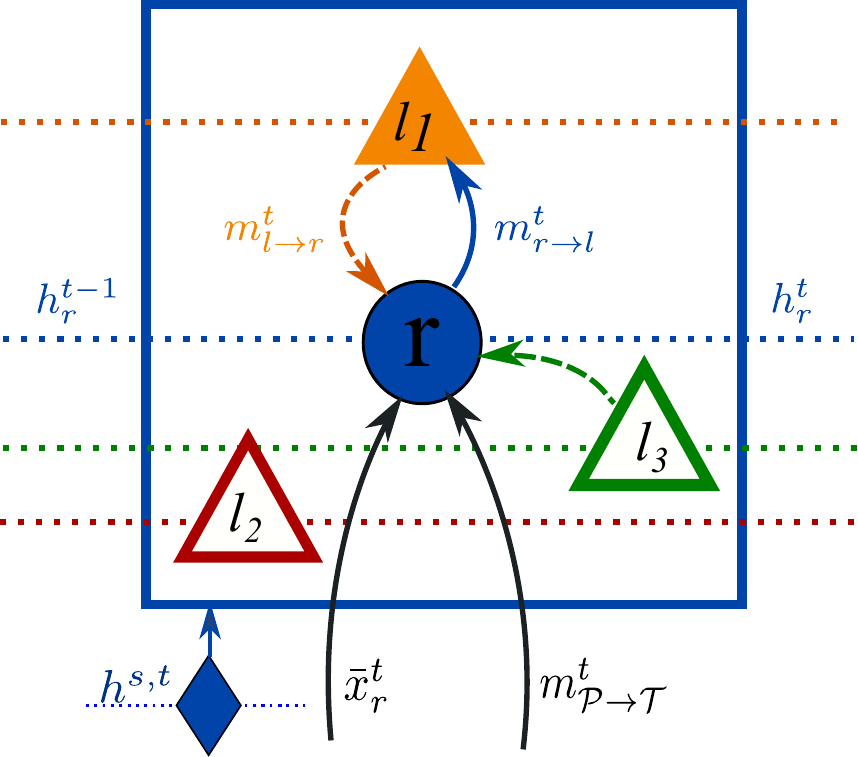}
\par\end{centering}

}\hfill{}\subfloat[Transient Switch\label{fig:transient-switch}]{\centering{}\includegraphics[width=0.47\columnwidth]{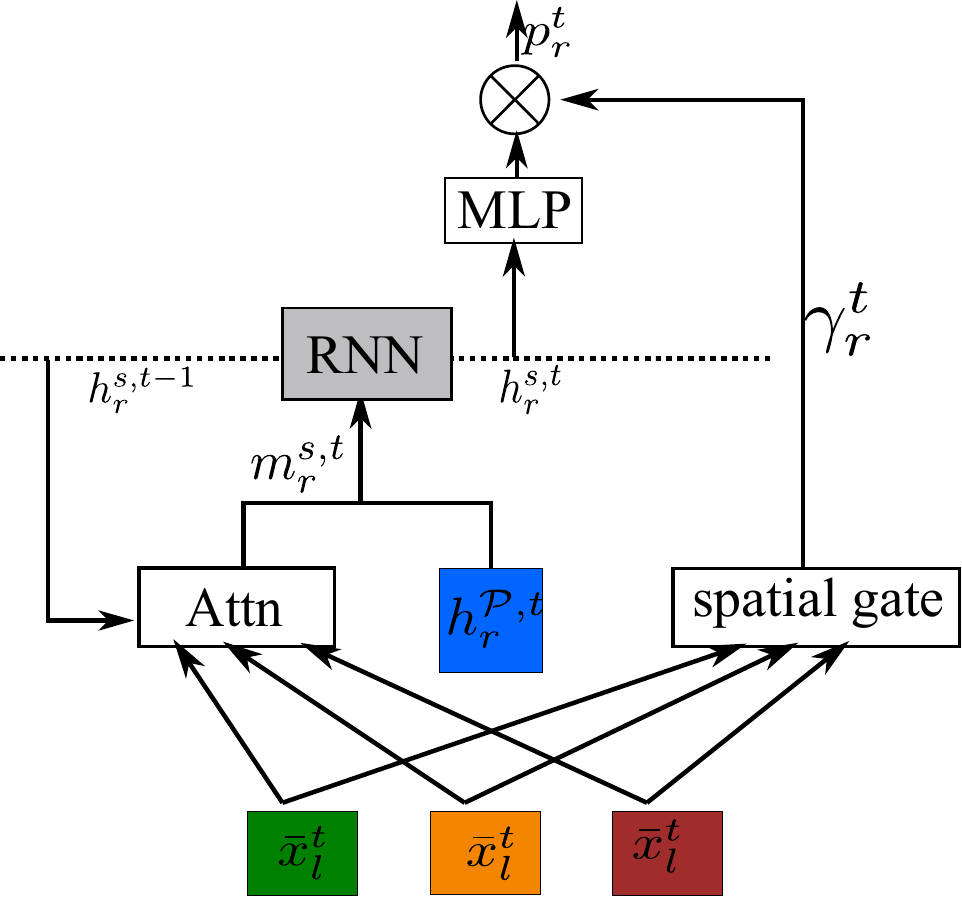}}

\caption{The Transient Channel (a) and Transient Switch (b)}
\end{figure}

\paragraph{The Egocentric representation}

of the entities are computed by transforming the geometrical features
into the egocentric coordinate system corresponding to the viewpoint
of the human center node $r$, $\bar{x}\_^{t}=f_{\text{ego}}\left(x\_^{t},x_{r}^{t}\right)=x\_^{t}-\text{centroid}(x_{r}^{t})$.
This change of system puts various patterns of the human's motion
into the same aligned space, filters out the irrelevant global information,
and facilitates efficient inference of the egocentric model.

\paragraph{The Egocentric inference}

is made by updating the RNN hidden state $h_{-}^{t}$ of each node
in the transient graph structure (see \ref{fig:transient-channel}).
In detail, for the \emph{center node}, the inward messages from its
leaves are aggregated into: $m_{l\rightarrow r}^{t}=\text{Attn}\left(\left[\bar{x}_{r}^{t};h_{r}^{t-1}\right],\left\{ \left[\bar{x}_{l}^{t};h_{l}^{t-1}\right]\right\} _{e_{l\rightarrow r}^{t}\in\mathcal{E}^{t}}\right)$.
It is then combined with the egocentric features $\bar{x}_{r}^{t}$
and the message from the persistent channel $m_{\mathcal{P}\rightarrow\mathcal{T}}^{t}$
(\ref{eq:persistent_output}) to update the RNN:
\begin{equation}
z_{r}^{t}=\left[\bar{x}_{r}^{t};m_{l\rightarrow r}^{t};m_{\mathcal{P}\rightarrow\mathcal{T}}^{t}\right],\hspace{0.3cm}h_{r}^{t}=\text{\text{RNN}}_{r}\left(z_{r}^{t},h_{r}^{t-1}\right).\label{eq:center-rnn}
\end{equation}

For\emph{ leaf nodes}, they only update their states if the center
node interacts with them, indicated by the outward edge

\vspace{-1em}

\begin{equation}
z_{l}^{t}=\left[\bar{x}_{l}^{t};m_{r\rightarrow l}^{t}\right],h_{l}^{t}=\begin{cases}
\text{\text{RNN}}_{l}\left(z_{l}^{t},h_{l}^{t-1}\right) & \textrm{if }e_{r\rightarrow l}^{t}\in\mathcal{E}^{t}\\
h_{l}^{t-1} & \text{otherwise}
\end{cases},\label{eq:local_leaf_hidden_states}
\end{equation}
where $m_{r\rightarrow l}^{t}$ is the outward message from the center
to its leaves, calculated from its hidden state through an MLP.

The updated hidden states are used generate the transient predictions
$\hat{y}_{i}^{\mathcal{T},t}$ and the messages sent to the persistent
process $m_{\mathcal{T}\rightarrow\mathcal{P}}^{t}$ (used in \ref{eq:persistent-rnn}):

\vspace{-1em}

\begin{align}
\hat{y}_{-}^{\mathcal{T},t} & =f_{\text{ego}}^{-1}\left(\text{MLP}\left(h_{-}^{t}\right)\right),m_{\mathcal{T}\rightarrow\mathcal{P}}^{t}=\phi\left(\text{MLP}\left(h_{r}^{t}\right)\right),\label{eq:transient-readout}
\end{align}
where $f_{\text{ego}}^{-1}$ is the function converting the egocentric
back to global coordinates. The transient predictions $\hat{y}_{-}^{\mathcal{T},t}$
are combined with those from the Persistent process in \ref{subsec:Future-prediction}.

\subsection{Switching Transient Processes \label{subsec:switch}}

The life cycles of the Transient processes are managed based on the
situation of human's activity by a neural \emph{Transient Switch}
(See \ref{fig:transient-switch}).

The switch first considers the current persistent state and the surrounding
environment of the center entity $r$ to update its switch RNN $h_{r}^{s,t}$:

\begin{align}
h_{r}^{s,t} & =\text{RNN}_{s}\left(\left[h_{r}^{\mathcal{P},t},m_{r}^{s,t}\right],h_{r}^{s,t-1}\right),\label{eq:switch_hidden}
\end{align}
where $m_{r}^{s,t}=\text{Attn}\left(h_{r}^{s,t-1},\left\{ \bar{x}_{l}^{t}\right\} _{l\neq r}\right)$,
and $\bar{x}_{l}^{t}=f_{\text{ego}}\left(x_{l}^{t},x_{r}^{t}\right)$
defined in \ref{subsec:transient_process}.

The RNN unit is important in maintaining the switch's sequential properties,
making it a state-full machine that can handle patterns of on and
off switchings and avoid spurious decisions caused by noises. The
switch-on probability $p_{r}^{t}$ is:

\begin{equation}
\hat{p}_{r}^{t}=\gamma_{r}^{t}\cdot\text{sigmoid}\left(Wh_{r}^{s,t}\right),\label{eq:switch_score}
\end{equation}
where $W$ is learnable weights. The discount factor $\gamma_{r}^{t}\in[0,1]$
responds to the distance from the subject to the nearest neighbor:
$\gamma_{r}^{t}=\exp\left(-\beta\cdot\text{min}\left\{ \left\Vert d_{lr}^{t}\right\Vert _{2}\right\} _{l\neq r}\right),$
where $d_{lr}^{t}$ are the center-leaf geometrical distances and
$\beta$ is a learnable decay rate. This factor acts as a disruptive
shortcut gate that modulates the switching decision based on the spatial
evidence of the interaction. 

Finally, the binary switch decision $\hat{s}_{r}^{t}$ is decided
by thresholding the switch score: the switch is on ($\hat{s}_{r}^{t}=1)$
when $\hat{p}_{r}^{t}>=0.5$, and is off otherwise. When it changes
from $0$ to $1$, a new Transient process is created at time $t$
for person $r$. This transient process will keep running until the
switch turns off, then the persistent process again becomes the single
operator.

\subsection{Future prediction\label{subsec:Future-prediction}}

In PTD, future motions are predicted by unrolling the model into the
future of $L$ time steps. At each future time step $t$, the predictions
from persistent channel $\hat{Y}^{\mathcal{P},t}$ (\ref{eq:persistent_output})
and those from Transient channel $\hat{Y}^{\mathcal{T},t}$ (\ref{eq:transient-readout})
are combined with the priority on the Transient predictions. For a
human entity, if its Transient channel is activated, the Transient
prediction will be chosen; otherwise, the Persistent prediction will
be used. For an object entity, if it receives an active outward Transient
edge, it will take that channel's prediction. If it receives multiple
outward edges, it uses the prediction from the channel with the highest
transient score $\hat{p}_{r}^{t}$. Otherwise, it uses the persistent
prediction by default.

\subsection{Model Training \label{subsec:losses}}

The model is trained end-to-end with two losses: prediction loss and
switch loss, $\mathcal{L}=\mathcal{L}_{\text{pred}}+\lambda\mathcal{L}_{\text{switch}}.$

The \textbf{Prediction loss} $\mathcal{L}_{\text{pred}}$ measures
the mismatch between predicted values $\hat{Y}$ and groundtruth $Y$,
$\mathcal{L}_{\text{pred}}=||\hat{Y}^{T:T+L}-Y^{T:T+L}||_{2}^{2}.$

The \textbf{Switch loss }$\mathcal{L}_{\text{switch}}$ is used to
supervise the Transient Switch and is implemented as: $\mathcal{L}_{\text{switch}}=\text{BCE}\left(\hat{P}^{1:T+L},P^{1:T+L}\right)$,
where $\hat{P}^{t}$ are the switch scores (\ref{eq:switch_score})
of all human entities, and $P^{t}$are binary ground-truth switch
scores collected at time step $t$.

\section{Experiment}

\begin{figure}[t]
\centering{}\includegraphics[width=0.95\columnwidth]{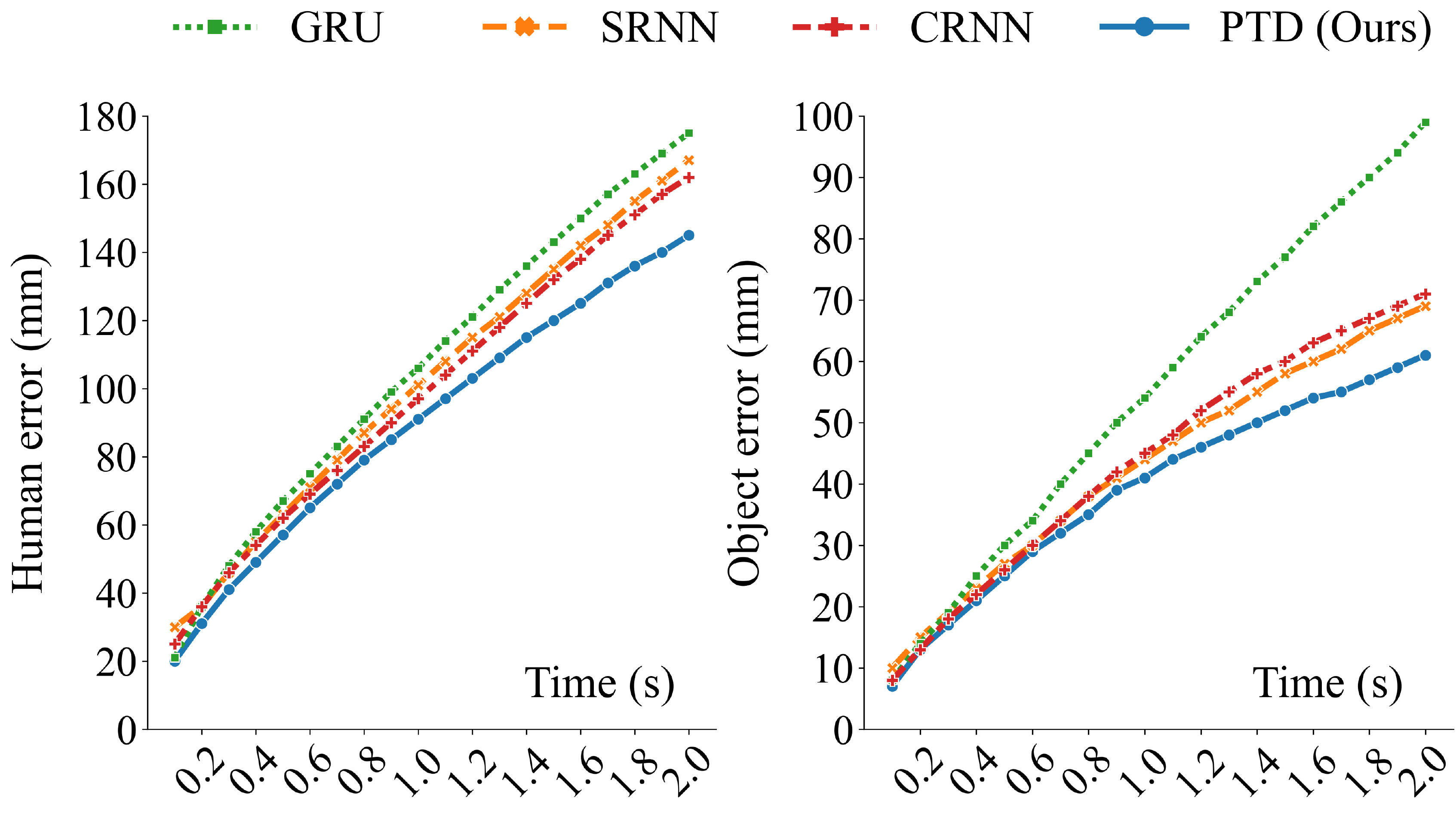}\caption{Quantitative performances and model size in HOI-M.\label{fig:qunatitative}}
\end{figure}

\paragraph{Experiment settings.}

Following the standard protocol \cite{corona2020context}, we extract
a subset of the Whole-Body Human Motion Database (WBHM) \cite{Mandery2015a},
which includes scenes that contain body poses of at least one human
entity, multiple movable objects, and a stationary object such as
``table''. This set includes 233 videos with 20 entity classes.
The selected features include 3D skeleton poses of 18 joints for human
entities and 3D bounding boxes for objects, sampled at 10Hz, consistent
with the compared methods \cite{corona2020context}.

We compare PTD with CRNN\cite{corona2020context}, Structural RNN\cite{jain2016structural},
and the simple GRU, whose implementations are redone for consistency.
We follow the common settings of observing 10 time steps (1 second)
and predicting future human and object motions for the next 20 time
steps (2 seconds). The models are conventionally trained on 80\% of
the videos and evaluated on the remaining 20\%. \vspace{-0.5em}

\paragraph{Quantitative evaluation.}

The performances are measured by prediction errors of joint positions
in the Euclidean distance (in mm). The means and standard deviations
from five independent runs are plotted in \ref{fig:qunatitative}
and show that PTD consistently outperforms the state-of-the-art, especially
in long-term prediction.

\begin{figure}[t]
\begin{centering}
\includegraphics[width=0.95\columnwidth]{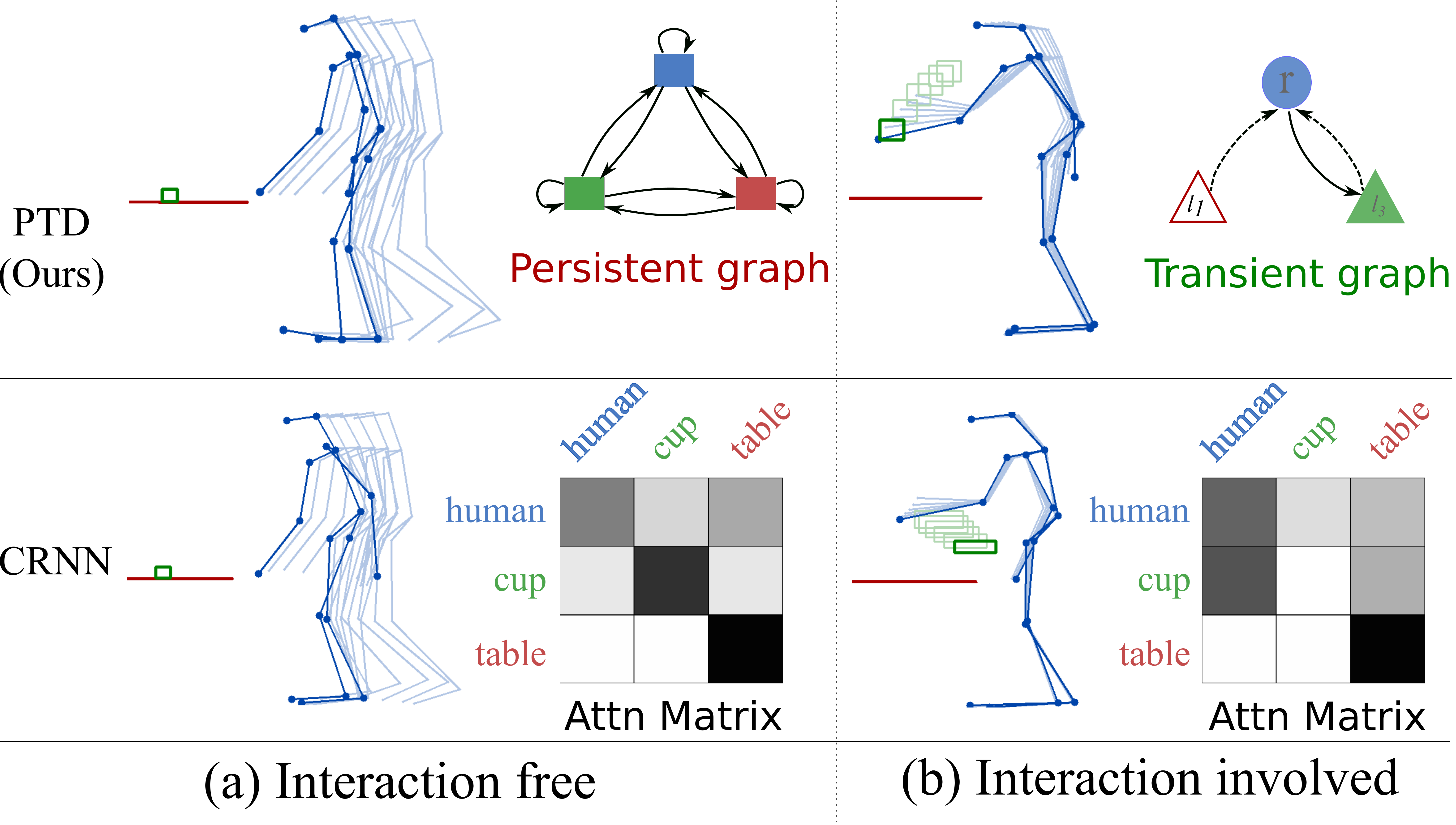}
\par\end{centering}
\caption{Persistent-transient duality in HOI-M. When the situation changes
from interaction-free (a) to interaction-involved (b), PTD \emph{(Upper
row}) switches on its Transient channel with egocentric structures
and handles the interaction accurately; In contrast, CRNN \cite{corona2020context}
(\emph{Lower row}) uses a single mechanism, resulting in the sluggish
adaptation of the attention map, leading to inaccurate predictions.
\label{fig:qualitative_hoi}}
\vspace{-0.5em}
\end{figure}

\paragraph{Visual analysis.}

We further verify the benefit of the duality by visualizing the internal
output predictions and graph structures of PTD compared to CRNN \cite{corona2020context}.
The upper row of \ref{fig:qualitative_hoi} shows that PTD could learn
to switch from the Persistent dense graph to the Transient egocentric
graph when the situation changes from interaction-free (a) to interaction-involved
(b). Particularly, in (b), the Transient graph reflects the interactions
correctly thanks to it being trained only on targeted interaction
samples free of noises.

In contrast, CRNN \cite{corona2020context} (lower row) holds on to
a single global mechanism and does not evolve adequately for the swift
change in the true relational topology, resulting in inaccurate and
unrealistic interactions.

\section{Conclusion}

In this work, we have introduced a new concept of the Persistent-Transient
duality to model the interleaving of global dynamics and the short-lived
interactions in human behavior. We model this conceptual duality into
a parent-child multi-channel network that can switch between the two
processes seamlessly. The superior performance of our model on the
HOI subset of WBHM confirms the effectiveness of this duality in human
behavior modeling.

\balance

{\small{}\bibliographystyle{ieee_fullname}
\bibliography{egbib}

\begin{thebibliography}{1}\itemsep=-1pt

\bibitem{baker2014fast}
Adam~P Baker, Matthew~J Brookes, Iead~A Rezek, Stephen~M Smith, Timothy
  Behrens, Penny J~Probert Smith, and Mark Woolrich.
\newblock Fast transient networks in spontaneous human brain activity.
\newblock {\em Elife}, 3:e01867, 2014.

\bibitem{corona2020context}
Enric Corona, Albert Pumarola, Guillem Alenya, and Francesc Moreno-Noguer.
\newblock Context-aware human motion prediction.
\newblock In {\em Proceedings of the IEEE/CVF Conference on Computer Vision and
  Pattern Recognition}, pages 6992--7001, 2020.

\bibitem{jain2016structural}
Ashesh Jain, Amir~R Zamir, Silvio Savarese, and Ashutosh Saxena.
\newblock Structural-rnn: Deep learning on spatio-temporal graphs.
\newblock In {\em Proceedings of the ieee conference on computer vision and
  pattern recognition}, pages 5308--5317, 2016.

\bibitem{Mandery2015a}
Christian Mandery, \"Omer Terlemez, Martin Do, Nikolaus Vahrenkamp, and Tamim
  Asfour.
\newblock The kit whole-body human motion database.
\newblock In {\em International Conference on Advanced Robotics (ICAR)}, pages
  329--336, 2015.

\end{thebibliography}
 }{\small\par}
\end{document}